\documentclass[conference]{IEEEtran}
\IEEEoverridecommandlockouts
\usepackage{cite}
\usepackage{amsmath,amssymb,amsfonts}
\usepackage{algorithmic}
\usepackage{graphicx}
\usepackage{textcomp}
\usepackage{xcolor}
\usepackage{todonotes}
\usepackage{url}

\def\BibTeX{{\rm B\kern-.05em{\sc i\kern-.025em b}\kern-.08em
    T\kern-.1667em\lower.7ex\hbox{E}\kern-.125emX}}

\definecolor{aoenglish}{rgb}{0.0, 0.5, 0.0}

\begin{document}

\title{Towards Game-Playing AI Benchmarks via Performance Reporting Standards}

\author{\IEEEauthorblockN{Vanessa Volz}
\IEEEauthorblockA{modl.ai, Denmark} 
\and
\IEEEauthorblockN{Boris Naujoks}
\IEEEauthorblockA{TH Cologne University, Germany}
}

\IEEEpubid{\begin{minipage}{\textwidth}\ \\[12pt]
978-1-7281-4533-4/20/\$31.00 \copyright 2020 IEEE
\end{minipage}}

\maketitle

\begin{abstract}
While games have been used extensively as milestones to evaluate game-playing AI, there exists no standardised framework for reporting the obtained observations. As a result, it remains difficult to draw general conclusions about the strengths and weaknesses of different game-playing AI algorithms. In this paper, we propose reporting guidelines for AI game-playing performance that, if followed, provide information suitable for unbiased comparisons between different AI approaches. The vision we describe is to build benchmarks and competitions based on such guidelines in order to be able to draw more general conclusions about the behaviour of different AI algorithms, as well as the types of challenges different games pose.
\end{abstract}

\section{Introduction}
\label{sec:Intro}

Artificial intelligence (AI) game-playing agents in commercial games are often used as in-game opponents for its players. As a result, such AIs 
need to fulfill several criteria.
A common criterion is how \textit{well} the agent can play the game -- this is significant for determining the challenge the player will encounter. For many types of games it is crucial that the AI agent embodies a particular role \textit{believably}. For example, a pathfinding algorithm may result in paths that a human would not normally follow. \textit{Efficiency} is another important criterion, as processing time is often at a premium in commercial games.

While these criteria are reflected in the evaluation of game-playing AI in some research-focused competitions, they are usually conducted on simplified games resulting in limited commercial interest. In contrast, several milestones of AI performance  against human players were achieved in popular complex games and widely covered in media. However, it is often difficult to draw general conclusions from these milestone performances. Adaptation into the games industry is also difficult due to dramatic computational costs.

What is lacking is a clear understanding of the strengths and weaknesses of AI algorithms and good approaches to predicting their approximate behaviour on a given complex game (or other AI challenge). However, this information is crucial for game developers who can only invest limited resources into exploration and research. It is also a cornerstone of research in academia, as better comprehension of an algorithm can inspire improvements and further research.

We argue that the adoption of general scientific standards for the evaluation and reporting on the performance of AI algorithms can help in gathering these missing insights. To this end, in this paper, we make the following main contributions:
\begin{itemize}
    \item A clear description of our vision of how benchmarks can be used to elicit empirically sound results that improve our ability to draw general conclusions about their behaviour and performance. This includes a discussion of potential issues with our approach and a list of requirements such benchmarks would need to fulfill.
    \item A set of guidelines intended as a starting point for a discussion of what information is required in order to allow the most efficient information gain for each set of empirical experiments conducted. In addition, we hope that reports adhering to such guidelines would lend themselves to unbiased and meaningful comparisons between two different game-playing AI agents. We propose to include (1) a detailed characterisation of the problem in order to identify behavioural patterns, (2) a description of the complexity of the AI solution in order to allow nuanced comparisons and (3) a specification of how results are collected and aggregated in order to record the setup and thus ensure their reproducibility as well as interpretability.
\end{itemize}

As stated above, there are often several different objectives that game-playing agents are developed to fulfil, including subjective aspects of their gameplay behaviour. \emph{Believability} is an important example of such objectives. While these subjective aspects are certainly very interesting and important, they are, by definition, difficult to measure automatically. They thus do not lend themselves well to automatic evaluation, especially if intended as larger scale empirical studies. As facilitating and encouraging such research is the main objective of this paper, we will focus on evaluating \emph{in-game performance} as measured by in-game score or winrates here.

In the following section, we first give some background on the evaluation of game-playing AI, including an overview of existing approaches, as well as more details on our motivation to propose benchmarking in this context. Next, we describe in more detail our vision of how benchmarks can be developed that are able to characterise general strengths and weaknesses displayed by tested AI algorithms, including a set of requirements for suitable benchmarks. As a first step towards fulfilling these requirements, we develop reporting guidelines for the performance of game-playing AI agents in section \ref{sec:ReportingGuidelines}. We conclude the paper in section \ref{sec:Conclusions} with a summary and suggestions for future work needed to implement the vision described in this paper.

\section{Background and Related Work}
\label{sec:Background}

Games are often cited as a promising testbed for AI, as problems are fully observable, flexible and reasonably complex to act as simulations for real-world processes \cite{gameaibook}. As games are designed to allow interaction between (often several) humans and a digital system, games are also well suited for researching these aspects of AI. As a result, there exist a plethora of environments for evaluating game-playing AI.

Below, we give definitions for several of these existing approaches of evaluating game-playing AI in order to allow a precise communication of our ideas. After that, we discuss related work for all of these approaches in more detail.

\paragraph*{Definition: Benchmarks} For the remainder of this paper, we explicitly define benchmarks to be sets of test cases for the given test subject (game-playing AI agents, in this case). Benchmarks thus evaluate different aspects of a given algorithm by posing a selection of different artificial challenges. Ideally, these tests cover all types of challenges expected to be encountered in the intended application.

\paragraph*{Definition: Milestone challenges} Milestone challenges are test case for algorithms (game-playing AI agents), but they are only made up of one specific problem setting that needs to be solved in a specific environment under specified constraints. Milestones thus constitute problems that push the boundaries of state-of-the-art research, but are narrowly defined.

\paragraph*{Definition: Competitions} While competition frameworks usually implement a collection of challenges and form a great basis for the implementation of a benchmark, the format of a competition does not lend itself well to drawing general conclusions. This is because by necessity, they need to define winning criteria, thus implicitly prioritising specific types of approaches and risking overfitting. This is a different intention from benchmarks, which seek to obtain an unbiased assessment of algorithms in different contexts without requiring a final ranking, following the \emph{no free lunch} principle.

\subsection{Human vs. AI Milestones}

Milestone-driven evaluations of AI have gained a large interest in the media in recent years. They are often conducted as events where an AI agent plays against human players of increasing skill level in a popular game. The winner is determined as specified in the game. 

Arguably one of the first milestone challenges was chess starting in the 1960's, gaining much publicity with the 1997 win of IBM's Deep Blue over the human chess champion Garry Kasparov \cite{ChessAI}. Increasingly complex games have been tackled since then, culminating recently in big news stories about AI beating human players in e.g. DotA 2 \cite{openai2019dota}. Real-time strategy (RTS) games, for example, have been proposed as an AI research challenge as early as 2003 \cite{Buro03real-timestrategy}, arguing that RTS games pose several AI research problems that were fundamental at that time. Since then, these games have been an area of active research \cite{SC2Survey}. Of course, Google DeepMind's AlphaStar reached Starcraft II Grandmaster level in 2019, playing in the game's European league against human players after winning several show matches against professional players\footnote{\url{https://deepmind.com/blog/article/alphastar-mastering-real-time-strategy-\\game-starcraft-ii}}. 

These wins of AI against skilled / professional human players are undoubtedly very impressive achievements and such challenges are continuing to motivate AI research. However, these events only generate very limited interpretable results which impedes the ability to draw reliable conclusions regarding the strengths and weaknesses of different AI approaches. Successful approaches in these challenges often consist of multiple interacting components, and require expensive computational resources as well as large datasets. Extensive tests are thus impractical and comparisons are impossible, as usually only a single solution exists. There further has been discussion on whether these matches are truly fair comparisons of in-game skill between human and AI players. \cite{Canaan2019} argue that at the time of writing, it is impossible to achieve perfect fairness across all dimensions. One example for the lack of fairness the authors state are the significant differences between the interfaces available to human and AI players.

\subsection{AI Research Competitions}

Besides matches against human players, many of the evaluation environments for game-playing agents published in research are used in context of competitions. The challenges and games range from pathfinding \cite{Sturtevant2014TheGP}, over Mario \cite{MarioAI} to StarCraft. Performance is usually evaluated against other AI systems. This can take the form of direct competition in a multiplayer game, such as the Bot Bowl competition for Bloodbowl, where competing AI systems play games against one another \cite{botbowl}. Competitions can also be indirect, e.g in single-player games that have easily comparable metrics, such as the Angry Birds competition which ranks AI systems based on how many points they score when playing the game \cite{angrybirds}.

In order to get more insight into the types of problems posed as competitions, we reviewed all competitions held at the IEEE Conference on Games 2019\footnote{\url{http://ieee-cog.org/2019/competitions_conference/}}. Most of the competitions use the in-game score as an evaluation measure for performance. However, some evaluate other aspects of AI players, such as generality (General Video Game Playing Competitions - GVGAI \cite{GVGAI}, ALE \cite{ALE}) or believability (2K Bot Prize in Unreal Tournament \cite{Hingston2009}). Here, we focus on in-game score and will therefore not be discussing these competitions further.

Further, in all surveyed competitions, the AI agent acted as a player in the game, as a human player would. This means no competitions targeted explicitly the development of AIs for non-player characters (NPCs), such as e.g. quest givers in a role-playing game. Still, overall, we find there are several different competitions that result in different challenges to AI agents. Unfortunately, each competition is evaluated separately and uses their own framework and execution environment. This makes comparisons of the same set of algorithms across different competitions impracticable or even infeasible. A given algorithm implementation with the same hyperparameter settings and evaluation resources is usually only evaluated in a single competition or by a single group of authors. 

At the same time, most competitions have a relatively narrow focus - often not broad enough to reach generalisable conclusions that could easily be applied to problems in the game industry. Still, due to the nature of competitions, even if the focus of a competition is perfectly aligned with a problem of interest, it remains difficult to generate transferable insights. This is because competitions usually only exist in specific settings and are not scalable. For example, a competition might require agents to make decision within a certain time-window. Results obtained from this setting do not necessarily contain information about the performance of agents when the allowed time for a decision is modified.

Summarising, we find that competitions are great drivers of research and a plethora of different challenges are tackled at the same time. However, it is also clear that there is a large amount of untapped potential for game-playing AI competitions that do not focus on in-game score. Further, while these competitions provide a number of very helpful evaluation environments, at the time of writing, their results can not easily be transferred to more general research insights.

\subsection{Interpretation of Results}

Some related work discusses the interpretation of AI evaluation results. This is mostly on a more theoretical basis, as usable empirical results are scarce. For example, in \cite{schaul2011measuring}, the authors make a case for using games to measure general intelligence. Various environments implement this idea, including the General Video Game Competition (GVGAI, \cite{GVGAI}) and the Arcade Learning Environment (ALE, \cite{ALE}). However, in \cite{chollet2019measure}, the author argues that state-of-the-art AI benchmarks is not currently capable of measuring general intelligence, as it is usually \emph{skill} exhibited at specific tasks that is measured instead.  

Other research has used data analysis to identify behavioural patterns in competition results \cite{horn}. However, an open issue remaining in these approaches is the incomplete understanding of the characteristics of the different problems the AI algorithms are tested on. The guidelines for reporting proposed in section \ref{sec:ReportingGuidelines} are intended to help alleviating this issue.

\section{Towards Interpretable Benchmarks}
\label{sec:InterpretableBenchmarks}
This section explains our vision of using benchmarks to collect interpretable empirical insights into the behaviour of game-playing AI algorithms. We first specify how we envision benchmarks to be used in this context and then develop a set of requirements needed to fulfil this vision.

\subsection{Motivation and Usecase}

Benchmarking of course can have various different purposes. However, one that is particularly relevant in the context of game-playing AI is identifying which AI approach should be used for a given problem. While it is often possible to make educated guesses about which approaches are promising, due to computational costs, it is often not feasible to test variations of algorithms thoroughly. This is especially problematic in complex games such as StarCraft II, where the only known well-performing AI (AlphaStar by Deepmind) combines various approaches and requires extensive amounts of computational resources to evaluate \cite{AlphaStar}.

Issues are especially apparent in the fact that, despite the widespread success of Monte-Carlo Tree Search (MCTS) over the last decade, with many examples of MCTS systems developed for competitions such as GVGAI\footnote{\url{http://gvgai.net/gvg_rankings.php}}, it has not seen a similar magnitude of impact on the games industry. One possible explanation for this is the cost of knowledge acquisition, and commercial games studios are often already operating on tight budgets in terms of both money and time. Even with clear explanations of how to implement a new algorithm, its benefits must be clear in order to justify the expenditure involved in adopting it. Our aim is to achieve industry-accepted standards by enabling meaningful comparisons between different approaches in relevant test environments.

We thus propose to develop benchmarks that primarily focus on being able to gain a better understanding of a given AI algorithm in the aforementioned usecase. Here, benchmarks can be useful in the following ways:
\begin{itemize}
    \item \emph{Understanding strengths / weaknesses of AI algorithms:} Gaining insight into which types of problems an AI has been proven successful at, is immensely helpful in identifying suitable candidate solutions for a given game.
    \item \emph{Measuring AI improvement:} Changing even small aspects of a game-playing AI often leads to significant behavioural changes due to the complexity of both AI and game environment\cite{ivanhyperparmeters}. Gathering information on behaviour changes to small AI adjustments in a benchmarking setting can help to identify promising modifications to test on the full scale.
\end{itemize}

\subsection{Identifying AI Abilities with Benchmarks}

Games pose complex problems to a player, independent of whether they are human or AI. The performance of a player on these problems thus gives indirect insight into their various abilities (such as e.g. the ability to react quickly, or to predict an opponent's behaviour). There is however no clear mapping between different abilities and a given game, as usually a combination of abilities is required. Thus, in order to gain useful information, we propose the following workflow:
\begin{enumerate}
    \item Identify research questions, i.e. which types of abilities are of interest in which context.
    \item Identify problems, i.e. tasks in games, where these abilities are required in different degrees. Ideally, it is possible to scale this requirement in the same game. This is relatively straightforward for abilities such as working memory, but might not be possible for others. Enough tasks need to be selected so that the abilities in question can be isolated, i.e. they are the only common ability required between the different problems.
    \item Describe the problems in detail, focusing on which abilities and cognitive skills are required, and what makes them difficult to solve for human and AI players. We discuss problem descriptions and characterisation more in section \ref{sec:Shaping}.
    \item Identify baseline performances in order to be able to interpret the results. Ideally, this includes the performance of some human players, as well as some basic AI players (random agent, as well as popular approaches such as Monte-Carlo Tree Search and Reinforcement Learning).
\end{enumerate}

\subsection{Benchmark Requirements}
\label{sec:requirements}

With our suggestion for the usage of benchmarks in mind, we have developed several criteria that a game-playing AI benchmark should fulfill. A benchmark should ...
\begin{itemize}
    \item produce measurable results. The target value should be meaningful and clearly defined.
    \item allow meaningful conclusions. The conclusions that can be drawn should be a meaningful resource of information for a research question / industrial games application.
    \item be interpretable. The results from the benchmark should allow conclusions on the behaviour of the tested AI under different conditions. This means a basic understanding of the included problems is necessary. It further needs to be established what the measured performance means in context of performance baselines, e.g. by observing random agents and average human players.
    \item produce generalisable conclusions. Results from benchmarks should ideally be transferable to similar games, algorithms, environments and hyperparameters. This can be achieved by facilitating the testing of different setups, such as different hyperparameter configurations and problems at different scales. Unfortunately, due to the complexity of the setup, this criterion is hard to guarantee. At least, known limits of generalisation should be disclosed.
    \item be reproducible. The same results should be reproducible during each run of the benchmark. If the results contain noise, it should be ensured that the benchmark is run often enough to produce robust statistics.
    \item produce robust results. Slight modifications to the game (e.g. modifying the colour of a game object) should not have a significant effect of the obtained results. This can be ensured by including multiple such modifications of the same setup in the benchmark and reporting averages.
    \item allow for comparisons. Two AI algorithms should be comparable based on their independently obtained benchmarking results without requiring further experiments.
    \item be practical. Running the benchmark and comparing to existing results cannot be prohibitively expensive considering commonly available computational resources. 
    \item disclose bias. Problems should be chosen with great care as to not favour specific algorithms or types of problems. Any consciously chosen bias should be disclosed.
    \item account for solution complexity. The complexity of each AI should be clearly communicated as it is not straightforward to compare algorithms with vastly differing computational costs.
\end{itemize}

Based on the above requirements, we identify three main types of information that needs to be reported in order to ensure their fulfilment:
\begin{itemize}
    \item Description of problems included in the benchmarks and the abilities required to solve them (cf. previous section).
    \item Description of the AI solution and its complexity.
    \item Details of how performance is quantified and aggregated, in particular including statistical information.
\end{itemize}

We develop our reporting guidelines in section \ref{sec:ReportingGuidelines} based on this conclusion.

\subsection{Limitations and a Word of Caution}
While benchmarks can undoubtedly help to identify the strengths and weaknesses of algorithms and thus inspire ideas for new algorithms or for improvements of existing ones, this type of results-driven research can also be fairly limiting for researchers. They should thus be used with caution.

For example, while we in this paper argue for a generalised reporting standard, we also want to stress that this standard needs to be adaptable enough to fit new or newly identified needs. While we aim for general applicability, it is clear that our proposal can and should be adapted to fit specific aims of different applications. We plan to give an explicit demonstration of such an adaptation in future work and modify the guidelines based on more detailed discussions. Furthermore, not all objectives for AI agents (such as believability) are easily quantifiable. Conventional performance-related benchmarks are thus not suitable as evaluation tools in these cases.

Even if great care is taken to represent a wide variety of challenges in a benchmark, ensuring appropriate coverage is difficult. Especially in the domain of games, even characterising a given test challenge is very problematic (see section \ref{sec:Shaping}). They should therefore be continuously evaluated for implicitly introduced biases and extended as much as possible. For the same reason, performance on a benchmark should never be used as the sole decision criterion for the success of an AI method or the acceptance of a publication. Only allowing submissions that improve some measurement, however well thought out, will inevitably lead to overfitting to this measure. This would most likely lead to an influx of papers about small modifications to existing algorithms that lead to equally small improvements in the measure specified. This would severely hinder progress in this field of research and must thus be avoided.

It is further important to acknowledge that for practical reasons benchmarking problems usually do not represent the full scale of their counterparts in the real-world. While this issue is unavoidable due to limits of computational resources, it is important to consider the additional challenges that arise through scaling up the size and complexity of a given problem. Consider for example the Kaggle Connect-X challenge\footnote{\url{https://www.kaggle.com/c/connectx}}, where agents are tasked to played versions of \emph{Connect 4} with a variable number of items that need to be connected ($X$). For small boards and a small number $X$, it is fairly straightforward to achieve game-theoretic perfect play with algorithms that exhaustively search the game tree. However, these approaches eventually run into issues with available computational resources when scaling up the problem. Technically, the type of problem hasn't changed, but the types of challenges resulting from it have. This is why defining the complexity of both the problem and AI solution (see section \ref{sec:requirements}) is important and the scalability of algorithms should be investigated.

Obviously, this raises the question: Why benchmark with simplified artificial functions at all? We argue that understanding algorithms and challenges in a detailed manner, even for smaller problems, can help divide and conquer larger-scale ones. Complex solutions rarely have end-to-end solutions and it is thus helpful to be able to understand the available building blocks and make educated guesses of how they could work together (see e.g. the many components working together in AlphaStar \cite{AlphaStar}. However, this also means that benchmarks do not offer all requirements for thoroughly evaluating a given game-playing agent. To advance the field, we should therefore also continue working on other approaches to evaluate AI performance, such as competitions and milestone challenges (see section \ref{sec:Background}).

\section{Reporting Guidelines}
\label{sec:ReportingGuidelines}

\begin{figure}
    \centering
    \includegraphics[width=\linewidth]{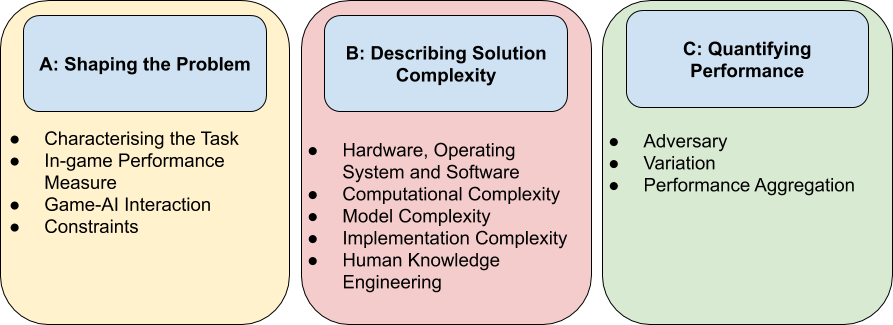}
    \caption{Proposed structure for reporting guidelines}
    \label{fig:structure}
\end{figure}

Based on the requirements we identified in the previous section, we concluded that in order to be able to draw general conclusions on the behaviour of AI agents, we need good reporting about (A) the problem, (B) the solution, and (C) the performance measurement procedure. Ideally, this reporting is standardised to some extent. A structure will help prevent the omission of facts that might not appear immediately relevant. Structure also facilitates a comparison between different algorithms, even if they were produced and tested in different contexts. The structure we use here is based on \cite{dagstuhl} and visualised in figure \ref{fig:structure}.

We propose such reporting guidelines below. They are intentionally kept relatively abstract to allow general applicability. However, this generality needs to be thoroughly tested in future work by applying these guidelines to several experiments.

\subsection{Shaping the Problem}
\label{sec:Shaping}

Several useful characteristics have been proposed to describe a problem, for example in \cite{gameaibook}. These include for example the number of players and the role the AI agent is assuming in the context of the game (e.g. an opponent, a teammate, a boss). However, these characteristics are not sufficient to identify the types of skills and abilities that would affect the performance of a player on a a given problem. We thus propose to describe the following additional aspects:
\begin{itemize}
    \item \textbf{Task}: What type of task the AI agent needs to perform (e.g. pathfinding), what types of obstacles exist that make this more difficult (e.g. dead ends) and abilities that would affect its performance (e.g. size of working memory).
    \item \textbf{In-Game Performance Measure}: How the degree of task fulfillment is reported (e.g. cost of the taken path)
    \item \textbf{Game-AI Interface}: How the AI agent interacts with the game and what information is made available (e.g. the game expects a cardinal direction from the agent)
    \item \textbf{Constraints}: Which constraints are set for the agent (e.g. maximum response time)
\end{itemize}

All of these aspects should be described in as much details as possible. Below, we propose a format for doing so in a structured manner.

\subsubsection{Characterising the Task}

Tasks in games often offer several different cognitive and physical challenges. Common examples of challenges posed to a player include decision making, resource management, planning, prioritisation, logical deduction, attentiveness, predicting an opponent's actions and quick reactions \cite{gameaibook}.

However, games usually do not challenge any specific ability separately and often introduce additional modifiers to vary the given problem. Let us look at chess as an example. Each turn, the player has to decide between a finite set of available legal moves. As all information is available, algorithms that search the gametree with minmax approaches seem suitable at first glance. However, due to the sheer size of the gametree caused by the relatively large branching factor, these approaches are not computationally feasible. If the branching factor was severely reduced, the type of the tasks would not change, but a big aspect of its challenge for an AI player would be removed.

Another example of added obstacles is hidden information. If some of the chess pieces were obscured by a \emph{fog of war} as is common in real-time strategy games, making a decision about the next move suddenly requires going through several possible gamestates and evaluating the options under uncertainty. Different games introduce different modifiers by e.g. introducing stochasticity or an adversary.

At the same time, abilities can be identified that are required to fulfill the challenges or to meet the modifications introduced by the added obstacles. In our chess example from above, the ability to process and store a larger amount of information (i.e. working memory) could help solve the problem introduced by a larger branching factor. Other examples of such abilities include reaction speed, timing, motor precision, perception and the ability to detect patterns in observed behaviour. These abilities are the information we are mainly interested in as a result from a benchmark.

Unfortunately, abilities do not map clearly to any given challenge or obstacle. For example, a player could react to hidden information by memorising previous states and predicting future game states based on these. An alternative way would be to consider uncertainty in their plans and develop a more robust strategy. A third option would be to just rely on quick reactions if new information is revealed unexpectedly. Most likely, it is best to improve all abilities. It is thus very important that the tasks are chosen and described with care.

For the time being, we can not provide a formal and exhaustive taxonomy that would facilitate the characterisation of a task. We suggest to approach this topic in collaboration with cognitive scientists. However, at this point in time, there is no consensus even in this field of research about how to classify different types of tasks and associated brain activity yet. While not in scope for this paper, in the future, we would like to be able to develop an exhaustive list of different types of tasks in games by conducting a large-scale survey.

\subsubsection{In-game Performance Measure}
\label{sec:ingame-perf}
It is of course important how the observed performance of the players is evaluated on the task as specified above, even if we focus exclusively on in-game measures. In particular, it is important to describe the manner in which players receive feedback on their measured in-game performance:
\begin{itemize}
    \item Timing: Does the player receive feedback continuously or only at the end of a round?
    \item Granularity: Does the performance measure reflect the degree of fulfilment of the task, or is it win / lose?
    \item Dependencies: Is the feedback absolute or relative, i.e. depending on other competing players?
\end{itemize}

The above aspects shape the problem faced by a player further. For example, late and scarce feedback might necessitate more exploration. Further, if performance is measured relative to other players, it is often helpful to react to other player's strength and weaknesses. The exact way how performance is measured and the manner in which the player receives feedback on it should thus be part of the problem description.

\subsubsection{Game-AI Interaction}

The way the interaction between the game and an AI algorithm is set up also shapes a large part of the overall problem. Through various competition frameworks, a defacto standard interface has been established in terms of the flow of information. The AI is usually expected to work within the game-loop, receiving information about the gamestate and additional resources if available at each tick. It is then expected to return an action which the game engine executes, resulting in the next gamestate. The agent performance is then measured and fed back to the player in a specific manner as discussed above.
Assuming this standard interface, the following aspects require description:

\paragraph{Game-State Representation} The representation of the game-state that is made available to the game-playing agent. For example, for a pathfinding task, sensible representations could be a graph or a grid. For playing arcade-style games, representations could be a list of sprites and their locations (like in GVGAI) or the game-output in pixels (in e.g. ALE). The type of representation plays a major role on what type of algorithms are suitable to the problem. Pixel representations require at least some form of image processing. Jump Point search, for example, only runs on grids, while A* runs on anything that can be represented as a graph, including grids.

\paragraph{Actions} There are several popular formats of actions, which are usually defined a priori. For example, in several cases, the interface expects an integer indicating which action should be chosen from a set of available actions (as in GVGAI). Another popular format is a location, along with an action and target unit (as e.g. in StarCraft II). It is further relevant to specify the number of available actions to choose from at each game tick (branching factor), as well as whether it is possible to choose invalid actions.

\paragraph{Additional Resources} Several setups also include additional information that is made available to the agent.
\begin{itemize}
    \item Forward model that simulates actions in the game. The availability of a forward model dictates whether statistical forward planning approaches are practical.
    \item Heuristics for gamestate evaluation. Heuristics can encode domain knowledge and can be immensely helpful to add performance feedback.
    \item Experimental setup. This allows extensive training on the specific setup, thus requiring less generalisation during testing. In some competition, the specific game used for evaluation is kept secret in order to encourage generality.
    \item Replays from human players. Replays can be used to bias the choice of suitable actions by modelling human behaviour. This can be especially helpful in environment with large action spaces.
\end{itemize}

\paragraph{Player Interaction} Another important aspect in this context is the manner of interaction with other players, especially in cooperative games. While some frameworks provide an additional communication interface, others do not allow communication at all (sometimes based on the game's rules).

\subsubsection{Constraints}

In some setups, constraints are specified for eligible solutions, which naturally further shape the challenge tackled. Common examples include the required reaction speed (40ms in GVGAI) or restrictions in terms of programming languages. We describe ways to describe the complexity of a solution in the following section. All of these aspects can also be used to define a constraint for valid solutions.

\subsection{Describing Solution Complexity}


As different complexities have a meaningful influence on the performance of game-playing AIs, reporting these complexities is important to allow for comparisons. Here we list several such complexities and explain these in a bit more detail. Of course, this list does not claim to be exhaustive.

\subsubsection{Hardware, Operating System and Software}
Core hardware components of a computer system (such as CPUs, GBUs and working memory) influence the execution of experiments significantly, especially regarding execution time. They should thus be part of any report on experiments. In addition, for the sake of reproducibility, we suggest to note and make available all software related to the experiments, including compilers, interpreters and libraries, as well as the operating system. 

\subsubsection{Computational Complexity}
Computational complexity describes the computational resources consumed during the execution of a given algorithm. Relevant numbers here include CPU and GPU workload, as well as consumed working memory and running time (both system and wall clock time). It is further important to indicate whether the resources were spent in a training phase or during online execution.

In addition, more domain-specific measures of computational complexity are relevant as well. For example how often a forward model was called or the response time of a given algorithm per round. In general, it is assumed that more expenditure of computational resources results in better algorithm performance.

	
	

\subsubsection{Model Complexity}
Model complexity describes the complexity of the algorithm after training, which relates both to its explainability, as well as to its specificity. It is often measured as a function of the number of hyperparameters and the architecture required to specify the the model. The more complex a model, the more prone it is to overfitting.





\subsubsection{Implementation Complexity}

The complexity of implementing a given algorithm is mainly relevant for its practicality. In addition, the less complicated an algorithm is, the easier it is to add modifications for further improvement.


\subsubsection{Human Knowledge Engineering}
Various parts of AI algorithms can be supported by inserting engineered domain knowledge and simultaneously reducing its generality. This reduces the complexity in other aspects of the algorithm, but requires formal domain knowledge instead. It should thus be carefully recorded what type of domain knowledge is integrated into the algorithm, specifically detailing inputs such as game state evaluation heuristics and constants chosen by domain experts. In addition, any domain-specific strategies and tactics the algorithm is based on should be disclosed.




\subsection{Quantifying Performance}

The in-game performance is of course defined based on the game (see section \ref{sec:ingame-perf}). However, the performance measure for an AI should depend on its intended purpose. For an AI designed as a team member in a cooperative game, for example, a suitable measure might be how closely its behaviour resembles human gameplay. If the intention is to produce agents that perform well in-game, naturally, measures such as the winrate or the average in-game score would be suitable. Of course, the more different measures are introduced, the more information is obtained on the AI performance.

Because most games and AI algorithms are non-deterministic, general standards for reporting the performance of empirical studies also apply here. One of the most important criteria in this regard is to communicate an assessment of the statistical significance of the obtained results. Further discussions on general best practices in the context of benchmarking can be found in other literature, such as \cite{eftimov2017novel}.

However, some fairly unique aspects of AI performance exist that need to be reported accordingly. For example, it is important to specify which \textbf{adversaries} the AI is tested against. We identified the following categories:
\begin{itemize}
    \item game: Single-player game where the game itself poses the challenge.
    \item NPC(s): Competitive multi-player game. Opponents are baseline AIs provided by environment (not competitors).
    \item game with NPC(s): Cooperative multi-player game. The game itself poses challenge, but is tackled together with baseline AIs provided by environment (not competitors).
    \item competitor AI(s): Competitive multi-player game. Opponents are other competing AI agents (past and present) as well as baseline AIs.
    \item humans: Human players.
\end{itemize}

For all setups that use match-ups against other submitted AIs, the resulting scores are by design relative to the other submissions. This is a common setup in competitions, as a ranking between all entries is desired. Further, in multi-player competitions, this tournament-style setup is used regularly, as in-game performance is usually not transitive. Well-known examples of this include sports tournaments and leagues. However, this style of competitions makes it difficult to add algorithms to the evaluation and the results are not interpretable out of context. They are thus not suited for evaluating a given algorithm independently, as is desired in benchmarks.

An important concern for evaluation is the robustness of the results. This is influenced both by the number of trials as well as the \textbf{variety} of test cases. It is difficult to make general statements about a suitable number of trials as this depends on the variance of obtained results. There are different sources of varied test cases:
\begin{itemize}
    \item inherent: Variety inherent in game and NPC behaviour. Basically fulfills the function of rematches and is intended to avoid drawing conclusions from statistical outliers.
    \item competitor(s): AIs are tested against competitors creating variety in how the opponent reacts to the agent's actions.
    \item instance: AIs are tested on different instances of the same game, e.g. different game maps or levels creating variety in the environment and the available actions per tick.
    \item games: AIs are tested on different games of the same genre. The environment, available actions and the effects of actions vary.
\end{itemize}

Many usecases of benchmarks, and especially competitions, require further \textbf{aggregation} of the different performance measures obtained this way. There are different popular approaches to this, depending on the setup:
\begin{itemize}
    \item Aggregated absolute performance: Mean or median of an absolute performance measure (e.g. single player games with in-game score as performance measure, or the winrate in multiplayer games against a immutable set of adversaries).
    \item Aggregated performance rank: In order to ensure equal weighting of each test case, some frameworks use the per-test-case-rank of each AI as a basis for their aggregated performance measure. GVGAI, for example, uses a Formula 1 scoring system per game and aggregates the allotted points for the final score. Another 6option is to perform Wilcoxon Rank-Sum tests on the obtained ranks.
    \item Aggregated relative performance: In multi-player games where the adversaries are not kept constant, the resulting performance measures per game are necessarily relative to the other AI agents. Popular ways of aggregating the results is by computing the win rates, by distributing points per match-up (as in many sports leagues) or by using iterative measures such as player rating (MMR in StarCraft II, ELO in chess).
\end{itemize}

As an AI player is designed to optimise the given performance measure, its specification of course defines an integral part of the problem it needs to solve. While current research, especially in competitions, focuses mostly on relative evaluations, we recommend to use absolute and usecase-specific measures for benchmarks so that algorithms are comparable independently. The adversaries and variety of testcases the measures are aggregated on should depend on the chosen usecase as well. This naturally requires the identification of a hypothesis prior to conducting a study. Further, it is important to consciously choose the different sets of testcases that the obtained performance measures are aggregated on, in order to avoid implicit weighting biases as well as interpretation errors.

\section{Conclusion and Future Work}
\label{sec:Conclusions}

In this paper, we detail how we envision benchmarks to be employed to understand the behaviour of game-playing artificial intelligence algorithms better. In particular, we are interested in characterising the abilities of algorithms and generalising these results across different games. A requirement crucial to our vision is the ability to report on different aspects of empirical results in detail, specifically the type of problems tested, the complexity of the AI algorithm in question, as well as how performance is quantified.

As a first step towards fulfilling these requirements, in this paper, we propose guidelines for reporting on empirical results of game-playing AI algorithms. In order to fully understand their usefulness, we plan to apply them to multiple games in the future. Based on resulting insights, we expect to be able to identify the usefulness and clarity of the different suggested reporting criteria. Following that, we plan to demonstrate the effectiveness of the obtained reports by conducting a detailed comparison between different AI algorithms, and identifying their respective strengths and weaknesses.

We also plan to identify more concepts from related disciplines of research, such as evolutionary optimisation, that could be helpful in the context of benchmarking game-playing AI. One promising example is the concept of anytime performance, i.e. measuring performance continuously for varying budgets. Another example is landscape analysis, where numerous low-level characteristics are computed to express specific aspects of a given problem. Data analysis approaches are then used to computationally identify patterns. In addition to a data-driven approach, other fields to consider are cognitive and neuro-science, especially when it comes to modelling mental challenges encountered in games.

\section*{Acknowledgements}

Boris Naujoks acknowledges the European Commission’s H2020 
programme, H2020-MSCA-ITN-2016 UTOPIAE (grant agreement No.\ 722734) 
as well as the DAAD (German Academic Exchange Service), 
Project-ID: 57515062 “Multi-objective Optimization for Artificial 
Intelligence Systems in Industry”.

Vanessa Volz acknowledges the helpful discussions on the topic of benchmarking during Dagstuhl Seminar 19511 with Yngvi Björnsson, Michael Buro, Mike Cook, Raluca D. Gaina, Günter Rudolph, Christoph Salge, Nathan Sturtevant, Tommy Thompson and Georgios N. Yannakakis.

\bibliographystyle{plain}
\bibliography{sample-base}

\end{document}